\documentclass[lettersize,journal]{IEEEtran}
\usepackage{amsmath,amsfonts}
\usepackage{algorithmic}
\usepackage{algorithm}
\usepackage{array}
\usepackage{textcomp}
\usepackage{stfloats}
\usepackage{url}
\usepackage{verbatim}
\usepackage{graphicx}
\usepackage{cite}
\usepackage{CJKutf8}
\usepackage{booktabs}
\usepackage{multirow}
\usepackage{siunitx}
\usepackage{subfigure}
\usepackage{bbding}

\begin{document}

\title{Dynamic Causal Disentanglement Model for Dialogue Emotion Detection}

\author{Yuting Su, Yichen Wei, Weizhi Nie\textsuperscript{*}, Sicheng Zhao, Anan Liu,~\IEEEmembership{Senior Member}~
\thanks{Yuting Su, Yichen Wei, Weizhi Nie, and Anan Liu are with the School of Electrical and Information Engineering, Tianjin University. Sicheng Zhao is with the School of Software, Tsinghua University. }
\thanks{Weizhi Nie is the Corresponding author, Email: weizhinie@tju.edu.cn.}}

\markboth{Journal of \LaTeX\ Class Files,~Vol.~14, No.~8, August~2021}%
{Shell \MakeLowercase{\textit{et al.}}: A Sample Article Using IEEEtran.cls for IEEE Journals}


\maketitle

\begin{abstract}
Emotion detection is a critical technology extensively employed in diverse fields. While the incorporation of commonsense knowledge has proven beneficial for existing emotion detection methods, dialogue-based emotion detection encounters numerous difficulties and challenges due to human agency and the variability of dialogue content.
In dialogues, human emotions tend to accumulate in bursts. However, they are often implicitly expressed. This implies that many genuine emotions remain concealed within a plethora of unrelated words and dialogues.
In this paper, we propose a Dynamic Causal Disentanglement Model based on hidden variable separation, which is founded on the separation of hidden variables. This model effectively decomposes the content of dialogues and investigates the temporal accumulation of emotions, thereby enabling more precise emotion recognition. 
First, we introduce a novel Causal Directed Acyclic Graph (DAG) to establish the correlation between hidden emotional information and other observed elements. Subsequently, our approach utilizes pre-extracted personal attributes and utterance topics as guiding factors for the distribution of hidden variables, aiming to separate irrelevant ones. Specifically, we propose a dynamic temporal disentanglement model to infer the propagation of utterances and hidden variables, enabling the accumulation of emotion-related information throughout the conversation. To guide this disentanglement process, we leverage the ChatGPT-4.0 and LSTM networks to extract utterance topics and personal attributes as observed information.
Finally, we test our approach on two popular datasets in dialogue emotion detection and relevant experimental results verified the model's superiority.
\end{abstract}

\begin{IEEEkeywords}
Emotion Detection; Structural Causal Model, Dynamic Causal Disentanglement Model
\end{IEEEkeywords}

\section{Introduction}\label{1}
With the ongoing advancements in computer technology and increasing technological sophistication, human-computer communication has spread to a wide range of applications. It is essential for robots to sense users' emotions and provide coherent, empathetic responses. This underscores the significant potential value of dialogue emotion detection across diverse fields. Emotion detection can be used effectively in recommendation systems, medical contexts, education, and beyond. With the proliferation of interactive machines, the importance of emotion detection is steadily increasing. Due to the wide spectrum of emotional expressions, particularly within varying topics and contexts, a single utterance may contain different emotional states, thus presenting a challenge for emotion detection.

With the exploration of NLP technology, researchers proposed some methods to solve emotional detection problems.
Khare et al. \cite{khare2020time} utilized CNN networks and proposed an emotion detection method based on time-frequency representation, using various convolutional neural networks for automatic feature extraction and classification.
Wang et al. \cite{wang2020wavelet} extracted emotion features using wavelet packet analysis from speech signals for speaker-independent emotion detection.
Jiao et al. \cite{jiao2020real} proposed an attention gated hierarchical memory network for real-time emotion detection processing.
Huang et al. \cite{huang2015bidirectional} proposed a Bi-LSTM network that can efficiently utilize past and future input features to classify.
Majumder et al. \cite{majumder2019dialoguernn} proposed a new method based on recurrent neural networks that keeps track of the individual party states throughout the conversation and utilizes this information for emotion classification.
Metallinou et al. \cite{metallinou2012context} focused on temporary emotional context and demonstrated the effectiveness of context-sensitive in emotion detection.

\begin{figure}{}
\centering         
\includegraphics[width=0.5\textwidth]{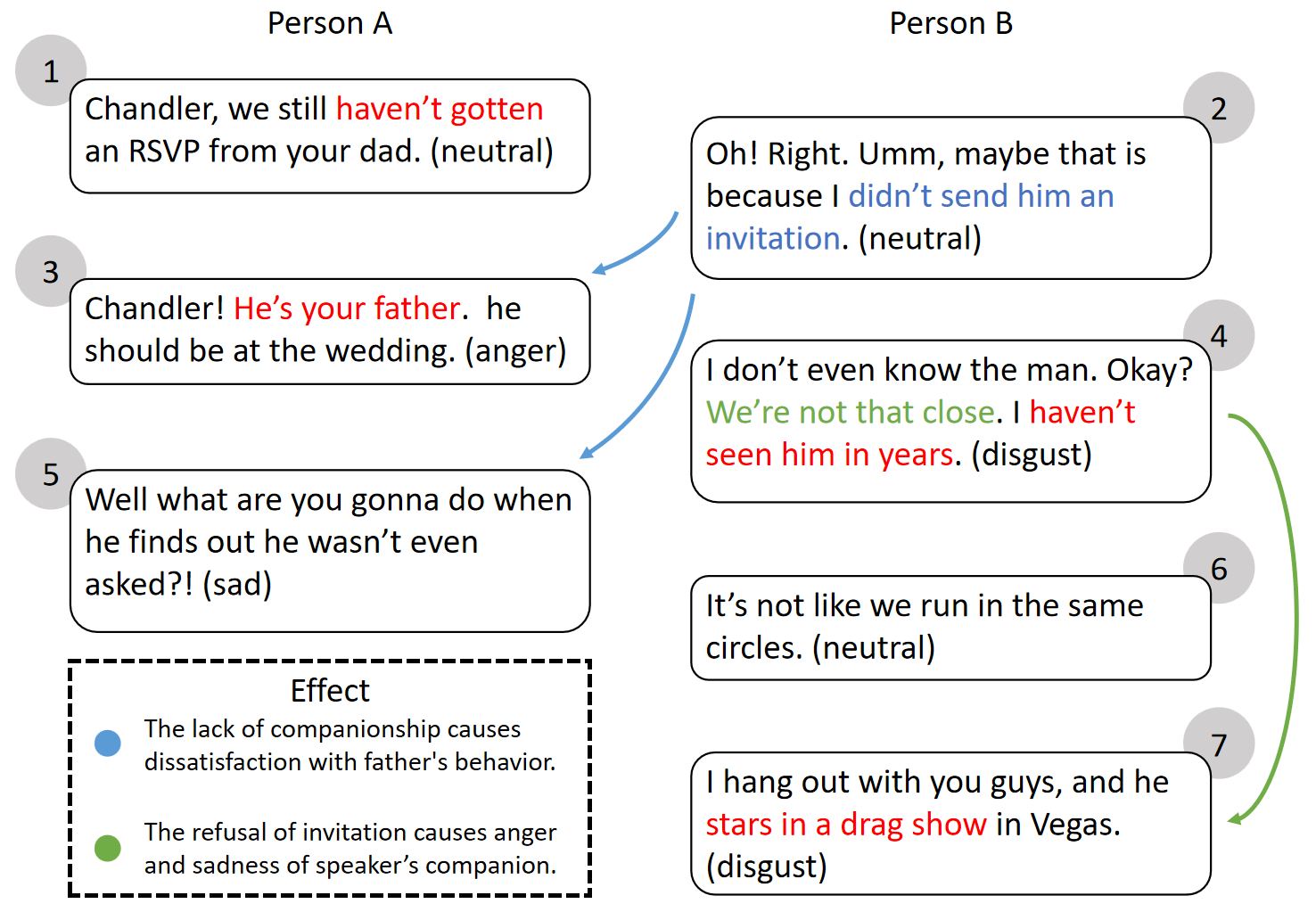}
\caption{The motivation diagram from the MELD dataset. Red-marked words are interfering words in the dialogue, whose meanings may mislead utterance emotion detection. Removing interference and using contextual information can better understand the speaker's psychology and emotions.}
\label{fig1}
\end{figure}

\subsection{\bf Motivation}
We find that prevailing techniques in dialogue emotion detection directly utilize the text of dialogue utterances for feature extraction and modeling. This approach, however, results in words unrelated to emotion in dialogues impeding accurate emotion detection. We view these unrelated words as spurious correlation information. Furthermore, we recognize that keywords in preceding utterances can exert influence on the emotion detection of subsequent utterances. Thus, acquiring contextual information from prior dialogues plays a pivotal role in guiding emotion detection.
As is shown in Fig.\ref{fig1}, it is a dialogue between two friends from the MELD dataset. We marked the interfering words in the dialogue utterances in red color. For example, “stars in a drag show” may mean a joyful and lively performance scene, but in fact, the emotion label of the utterance is \textit{Sadness}. We find the distant relationship mentioned in the previous dialogue can better explain the speaker's sadness in this utterance. We posit that the utilization of the previous context and the exclusion of spurious correlation information in utterances can significantly improve the accuracy of emotion detection.

In our work, to avoid the impact of spurious correlation information, we propose a Causal Disentanglement Model. Our objective is to model hidden variables which propagate to generate the utterances, and to separate hidden variables unrelated to emotions from other variables. The causal relationship is illustrated in Fig.\ref{fig2}
Among the hidden variables constituting the utterance $U$, namely $s$, $v$, and $z$, we designate $s$ as the hidden variable associated with emotions and contributing to utterance composition, $v$ as the hidden variable linked to dialogue topics related to emotions, and $z$ as the hidden variable unrelated to emotions. As depicted in the figure, both $s$ and $v$ have a causal relationship with the emotion $E$, while $v$ possesses a causal relationship with the dialogue topic $F$. We apply ChatGPT to complete the text feature extraction task, i.e., extraction of the dialogue topic $F$. Subsequently, we augment the dataset with the extracted topic text and topic feature vectors for future research. Notably, throughout each time stage, the evolution and generation of all hidden variables are influenced by the auxiliary variable $P$, which represents personal attribute information derived from previous conversations using an LSTM network. In addition, we propose a sequential VAE framework \cite{kingma2013auto} to reconstruct dialogue utterance $U$ and dialogue topic $F$ as part of our model optimization process.


\subsection{\bf Contributions}
This paper makes the following main contributions:
\begin{itemize}
    \item[\textbullet] We propose a Dynamic Causal Disentanglement Model, which aims to separate hidden variables unrelated to emotions from dialogue utterances. Our model significantly improves the robustness of emotional features.
    \item[\textbullet] We apply ChatGPT to complete the feature extraction task and add the topic text and feature vectors to the existing dataset.
    \item[\textbullet] We evaluate the performance of our method on the IEMOCAP and MELD datasets. The experimental results indicate that our method has significant improvements compared to the most advanced methods.
\end{itemize}

In Section \ref{2}, we present the relevant work. In Section \ref{3}, we briefly introduce the background of the causal model. In Section \ref{4}, we introduce the specific content of the model and the calculation formula of the methodology in detail. In Section \ref{5}, we present key experiments. In Section \ref{6}, we present the conclusions of our work and discuss future research.

\section{Related Work}\label{2}
\subsection{\bf Emotion detection}
Emotion detection represents a prominent and actively researched subfield within natural language processing, finding widespread application in diverse domains, including the medical field, intelligent devices, and the Internet of Things \cite{zhu2023uaed, li2023global, kim2020virtual, ding2020tsception}. The initial research on dialogue emotion detection is mainly based on audio features and video features \cite{datcu2015semantic}. Recently, the development of deep learning drives the research on dialogue emotion detection in text \cite{singh2023text} and speech \cite{bhattacharya2022deep}.

Long Short Term Memory (LSTM) networks \cite{hochreiter1997long} is a temporal recurrent neural network. It is one of the most common and effective methods for dealing with sequential tasks.
Poria et al. \cite{poria2017context} proposed a detection model-based LSTM that enables discourse to capture contextual information from the surrounding environment within the same video.
Li et al. \cite{li2020exploring} proposed a multimodal attention-based BLSTM network framework for detection. They propose Attention-based Bidirectional Long Short-Term Memory Recurrent Neural Networks (LSTM-RNNs) to automatically learn the best temporal features that are useful for detection.
Huang et al. \cite{huang2019ana} propose a novel Hierarchical LSTMs for Contextual Emotion Detection (HRLCE) model.

In recent studies, the incorporation of common sense knowledge has been increasingly prevalent. In tasks involving inference, the integration of common sense knowledge facilitates the deduction of implicit information from textual content.
Ghosal et al. \cite{ghosal2020cosmic} proposed COSMIC, a new framework that includes different commonsense elements to assist in detection.
Zhong et al. \cite{zhong2019knowledge} leveraged commonsense knowledge to enrich the transformer encoder.
Li et al. \cite{li2021enhancing} proposed a conversation modeling module to accumulate information from the conversation and proposed a knowledge integration strategy to integrate the conversation-related commonsense knowledge generated from the event-based knowledge graph.
Bosselut et al. \cite{bosselut2019comet} proposed commonsense transformers (COMET) which are used to learn to generate rich and diverse commonsense descriptions in natural language.

Graph models are widely applied to process and analyze semantics and relations in the texts. The latest progress in graph model research has also been introduced into the field of emotion detection.
RGAT \cite{ishiwatari2020relation} can capture the speaker dependency and the sequential information with the relational position encodings which provide the RGAT model with sequential information reﬂecting the relational graph structure. IGCN \cite{nie2021gcn} applied the incremental graph structure to imitate the process of dynamic conversation.
Choi et al. \cite{choi2021residual} utilized the residual network (ResNet)-based, intrautterance feature extractor and the GCN-based, interutterance feature extractor to fully exploit the intra-inter informative features.

Nevertheless, these approaches fail to disentangle features causally related to emotions from dialogue utterances, which results in spurious correlations and diminished robustness in feature extraction.

\subsection{\bf Causal Hidden Markov Model}
The Causal Hidden Markov Model is a common statistical model, which is widely utilized in many time series problems, including image processing, natural language processing, bioinformatics, and other fields. The Causal Hidden Markov Model is an extension of the Markov process that integrates hidden states. 
In contrast to the standard Markov Model, the Causal Hidden Markov Model excels in capturing the causal relationship between observation sequences and hidden state sequences by acquiring knowledge of the transition probabilities and probability distributions of observations and hidden states. This enhanced capability contributes to a more effective modeling of time series data.
Many methods are proposed based on Causal Hidden Markov Models.
Khiatani et al. \cite{khiatani2017weather} constructed the Hidden Markov Model to learn the relationship between weather states and hidden state sequences, achieving more accurate weather prediction.
ZHANG et al. \cite{zhang2020non} utilized the Factorial Hidden Markov Model Based on the Gaussian Mixture Model for non-invasive load detection.
Guo et al. \cite{guo2021predictive} attempted to accurately predict the chloride ion concentration that causes bridge degradation, and in the face of random detection time intervals, proposed a predictive Hidden semi-Markov Model to achieve prediction.
Zhao et al. \cite{zhao2023causal} proposed a Causal Conditional Hidden Markov Model to predict multimodal traffic flow, which respectively designs a prior network and a posterior network to mine the causal relation in the hidden variables inference stage.
Mak et al. \cite{mak2011causal} proposed a Causal Topology Design Hidden Markov Model to solve viewpoint variation issue, which can help view independent multiple silhouette posture recognition.
Suphalakshmi et al. \cite{suphalakshmi2009full} utilized a Full Causal Two Dimensional Hidden Markov Model with a novel 2D Viterbi algorithm for image segmentation and classification.

\begin{figure*}   
\centering         
\includegraphics[width=18cm]{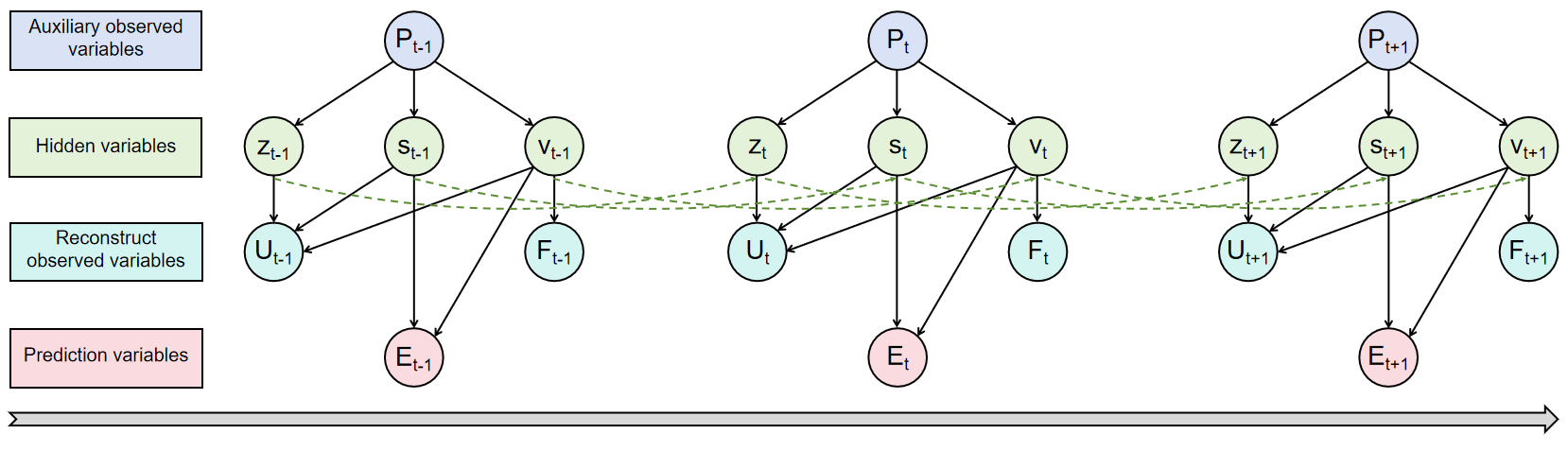}
\caption{Directed acyclic graph for emotion detection. $s$, $v$, and $z$ represent hidden variables associated with emotions and utterance composition, dialogue topics related to emotions, and variables unrelated to emotions, respectively. Personal attributes $P_t$ influence the distribution of the hidden variables $s_t$, $v_t$, and $z_t$, which collectively form the utterances $U_t$. Following disentanglement, we reconstruct both the utterances $U_t$ and the topics $F_t$ for optimization purposes. At each time step, $s_t$ and $v_t$ are employed to derive the emotion label $E_t$.}
\label{fig2}
\end{figure*}

\section{Preliminaries}\label{3}
We commence with an introduction to Structural Causal Models (SCMs). The SCM refers to a causal graph represented by a directed acyclic graph associated with structural equations. The causal graph is denoted as $G:= (V, E)$, where $V$ and $E$ represent node set and edge set respectively. In the edge set,  each arrow $x \rightarrow y$ $(x, y \in E)$ indicates that $x$ has a direct influence on $y$ and changes the distribution of $y$. The structural equation contains the production relation of each variable in the point set. 
In the node-set, for $V:= \{v_1, \ldots, v_k\}$, we define $p(v_i)$ to represent the set of parent nodes of $v_i$, and the set of the parent nodes will affect it. Through the production relation, we define the relation as $v_i \leftarrow f_i(p(v_i))$ where $f_i$ denotes causal mechanisms.

\section{Methodology}\label{4}
Inspired by the Causal Hidden Markov Model \cite{li2021causal}, we propose the Dynamic Causal Disentanglement Model to model dialogues as a novel approach for modeling dialogues. Our modifications and enhancements to the model are tailored to facilitate the detection of emotional content within dialogue utterances. In our Dynamic Causal Disentanglement Model, we denote observed variable $f_{t} \in F$, $u_{t} \in U$, $p_{t} \in P$ as the dialogue topics, dialogue utterances, and personal attributes at time stage t. The $e_{t} \in E$ is the predicted emotion label at time stage t. We disentangle the utterance into hidden variables $s_{t}$, $v_{t}$, $z_{t}$. 
To observe the progress of dialogues clearly, we design an LSTM network to learn the features of the speaker's attributes at the current time stage. Due to the recent popularity of ChatGPT, we apply ChatGPT-4.0 to extract dialogue topic features. 

In this section, we introduce the Dynamic Causal Disentanglement Model in detail and elucidate the methods used for learning hidden variables within the model.

\begin{figure*}  
\centering         
\includegraphics[width=16cm]{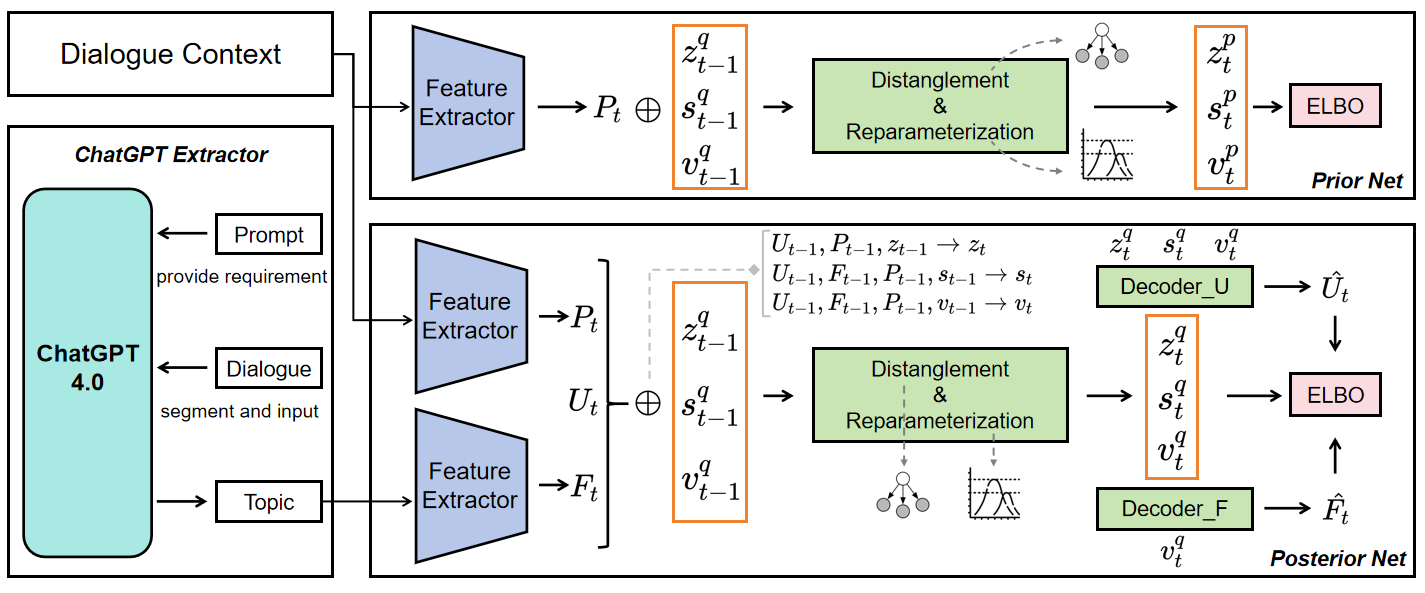}
\caption{The network structure of Dynamic Causal Disentanglement Model. In causal learning, we propose prior and posterior networks to learn the distribution of variables to achieve disentanglement. In the prior network, we input the concatenated vectors into the GRU and subsequently design two independent Fully Connected (FC) layers to obtain the mean and the log variance vectors. Within the posterior network, we input concatenated vectors into the FC layer, similar to the prior network, to obtain mean and log variance vectors. The hidden variable data obtained from the posterior network is then utilized for the reconstruction of observed variables and the prediction of labels.}
\label{fig3}
\end{figure*}

\subsection{\bf Dynamic Causal Disentanglement Model}
To describe the dialogue progresses, we introduce the causal directed acyclic graph (CDAG) shown in Fig.\ref{fig2}. In the CDAG, the causal relation of variable $m$ is formulated as $f_m := (p_a(m), \varepsilon_m)$, where $p_a(m)$ is parent nodes of node $m$ and $\varepsilon_m$ is independent exogenous variables. All variables form a Hidden Markov model.
We define the model as follows: 
\begin{equation*}
\begin{aligned}
& z_t \leftarrow f_{z}(P_{t}, \varepsilon_{z}^{t}), v_t \leftarrow f_{v}( P_{t}, \varepsilon_{v}^{t}), s_t \leftarrow f_{s}(P_{t}, \varepsilon_{s}^{t}), \\
& P_{t} \leftarrow f_{P}(\varepsilon_{P}^{t}), F_t \leftarrow f_{F}(v_t, \varepsilon_{F}^{t}), E_t \leftarrow f_{E}(s_t, v_t, \varepsilon_{E}^{t}), \\
& U_t \leftarrow f_{U}(s_t, v_t, z_t, \varepsilon_{U}^{t}).
\end{aligned}
\end{equation*}

To enhance detection accuracy, our objective is to disentangle emotion-related information. We introduce hidden variables $s_{t}$, $v_{t}$, and $z_{t}$ to construct the observation variables $U_{t}$, $F_{t}$, and $E_{t}$ at respective time stages. Concurrently, the hidden variables are influenced by the auxiliary variable $P_{t}$, which undergoes continuous updates throughout the ongoing dialogue. $s_{t}$ represents the hidden variable associated with emotions and contributing to utterance composition, $v_{t}$ represents the hidden variable linked to dialogue topics related to emotions and $z_{t}$ represents the hidden variable unrelated to emotions. Collectively, these hidden variables form a comprehensive utterance representation. The hidden variables $s_{t}$ and $v_{t}$, which impact emotional assessment, are related to the emotion label $E_{t}$, whereas the hidden variable $v_{t}$ pertaining to the dialogue topic solely influences the dialogue topic $F_{t}$. Improved predictive accuracy can be achieved by isolating the hidden variables contributing to the utterance and eliminating extraneous variables. For simplicity, we define $h := \{s, v, z\}$ ; $o := \{P, F, U\}$. According to Causal Markov Condition \cite{pearl2009causality}, we calculate the factorization of joint distribution as:
\begin{equation}
\begin{gathered}
p(h_{\leq T}, o_{\leq T}, E_{\leq T}) = \prod_{t=1}^{T} \biggl[ p(E_t | s_t, v_t) \cdot
\\
p(h_t | h_{t-1}, P_{t}) \cdot p(U_t | h_t) \cdot p(F_t | v_t) \biggr].
\label{eq1}
\end{gathered}
\end{equation}

We aim to analyze the distribution of hidden variables from the prior distribution $p_\psi (h_t | h_{t-1}, P_t)$. Due to the difficulty of calculation, we choose to utilize posterior distribution $q_{\phi}(h_{\leq T} | o_{\leq T}, E_{\leq T})$ for approximation \cite{blei2017variational}. $p(E_t | s_t, v_t)$, $p(U_t | h_t)$, $p(F_t | v_t)$ are the processing of learned hidden variables in the generation network. After defining the joint probability distribution, we propose a sequential VAE framework to learn our causal model, as shown in Fig.\ref{fig3}.

\subsubsection{\bf \textit{Prior Network}}
For establishing prior distributions, we introduce a prior network into our model. This prior network takes both the observed variable $P$ and the hidden variables from the preceding time step as inputs. We utilize Gated Recurrent Units (GRUs) \cite{chung2014empirical} to propagate hidden variables, enhancing the model's capacity to capture long-term dependencies. After the GRUs unit, we design independent Fully Connected layers (FCs) to obtain the mean and the log variance vectors of the hidden variables. We utilize the hidden variable $h \in \{s, v, z\}$ to illustrate the disentanglement propagation process of GRUs as follows:
\begin{equation}
\begin{gathered}
r_t^p = \sigma(W_r^p \cdot [h_{t-1}, P_t] + b_r^p),\\
k_t^p = \sigma(W_z^p \cdot [h_{t-1}, P_t] + b_k^p),\\
\tilde{h}_t^p = \tanh(W^p \cdot [r_t \odot h_{t-1}, P_t] + b^p),\\
h_t^p = (1 - k_t^p) \odot h_{t-1}^p + k_t^p \odot \tilde{h}_t^p.
\label{eq2}
\end{gathered}
\end{equation}

\noindent $r_t^p$ and $z_t^p$ represent the status of the reset gate and update gate. Hidden update status $h_t^p$ is determined by candidate hidden status $\tilde{h}_t^p$ and update gate status $z_t^p$. In the above formula, $\sigma$ represents the sigmoid function, and $\odot$ represents the element-wise multiplication operation. All hidden variables are learned in the same method.

\subsubsection{\bf \textit{Posterior Network}}
In probability graph models, solving the posterior distribution of hidden variables through Bayesian formulas can be challenging. Consequently, we opt for using a known simple distribution to approximate the intricate distribution that requires inference. We utilize distinct units to learn posterior distributions as follows:
\begin{equation}
\begin{gathered}
s_t^q = W_s^q \cdot [s_{t-1}, U_t, F_t, P_t] + b_s^q,\\
v_t^q = W_v^q \cdot [v_{t-1}, U_t, F_t, P_t] + b_v^q,\\
z_t^q = W_z^q \cdot [s_{t-1}, U_t, P_t] + b_s^q.
\label{eq3}
\end{gathered}
\end{equation}

\noindent Similar to prior networks, we apply Fully Connected layers (FCs) to obtain the mean and the log variance vectors of the hidden variables. These vectors will be utilized for reconstruction in the following section.

\subsubsection{\bf \textit{Generation Network}}
In the generative network, we employ the updated hidden variables to reconstruct the observed variables and predict emotion labels. To guarantee the representation ability of learned variables, we utilize $s_t^q$, $v_t^q$, $z_t^q$ to obtain the reconstructed utterance $\hat{U}$, and $v_t^q$ to obtain the reconstructed conversation topic $\hat{F}$. We optimize our model by approximating the reconstructed variables $\hat{U}$ and $\hat{F}$ to the observed variables $U$ and $F$.

\subsection{\bf  Learning Method}
The optimization target is the Evidence Lower Bound (ELBO). ELBO consists of evidence and KL divergence and maximizing ELBO is tantamount to minimizing the KL divergence, intending to approximate the posterior distribution as closely as possible to the true posterior distribution.
ELBO is represented as:
\begin{equation}
\begin{gathered}
\mathbb{E}_{p(o_{\leq T}, E_{\leq T})} \Big[
E_{q_{\phi}(h_{\leq T} | o_{\leq T}, E_{\leq T})} \log\left(p_{\psi}(o_{\leq T}, E_{\leq T})\right)
\\
-\mathcal{D} \left(q_{\phi}(h_{\leq T} | o_{\leq T}, E_{\leq T})  \parallel p_{\psi}(h_{\leq T} | o_{\leq T}, E_{\leq T}) \right) \Big],
\label{eq4}
\end{gathered}
\end{equation}

\noindent where $\mathcal{D} (\cdot \parallel \cdot)$ denotes KL divergence. After simplifying the evidence and KL divergence, we obtain:
\begin{equation}
\begin{gathered}
\mathbb{E}_{p(o_{\leq T}, E_{\leq T})} \Big[ E_{q_{\phi}(h_{\leq T} | o_{\leq T}, E_{\leq T})} \log\left(\frac{p_{\psi}(h_{\leq T}, o_{\leq T}, E_{\leq T})}{q_{\phi}(h_{\leq T} | o_{\leq T}, E_{\leq T})}\right) \Big].
\label{eq5}
\end{gathered}
\end{equation}

According to Eq.\ref{eq1}, we know the prior distribution $p_{\psi}$. We consider the mean and the log variance vectors in the prior network as the learned results, $p_{\psi}(\alpha_t|\alpha_{t-1}, P_{t})$ for each $t$ is distributed as $\mathcal{N}(\mu_{\psi}(\alpha_{t-1}, P_{t}), \Sigma_{\psi}(\alpha_{t-1}, P_{t}))$. In the posterior distribution, the posterior is given by:
\begin{equation}
q_{\phi}(h_{\leq T} | o_{\leq T}, E_{\leq T}) = \frac{\displaystyle\prod_{t=1}^{T} \left[ q_{\phi}(E_t | s_t, v_t) \cdot q_{\phi}(h_t | h_{t-1}, o_t) \right]}{q_{\phi}(E_{\leq T} | o_{\leq T})}.
\label{eq6}
\end{equation}

Under the similar reparameterization, $q_{\phi}(h_t | h_{t-1}, o_t)$ for each $t$ is distributed as $\mathcal{N}(\mu(h_{t-1}, o_t), \Sigma(h_{t-1}, o_t))$. Substituting Eq.\ref{eq1} and Eq.\ref{eq6} in Eq.\ref{eq5}, we reformulate the ELBO as:
\begin{equation}
\mathbb{E}_{p(o_{\leq T}, E_{\leq T})} \left[ \log q_{\phi}(E_{\leq t} | o_{\leq T}) + \mathcal{L}_{q_{\phi}, p_{\psi}} \right].
\label{eq7}
\end{equation}

\noindent In Eq.\ref{eq7}, following the derivation based on probability theory, we know that $q_{\phi}(E_{\leq T} | o_{\leq T})$:
\begin{equation}
\int \prod_{t=1}^{T} [q_{\phi}(E_t | s_t, v_t) \cdot q_{\phi}(h_t | h_{t-1}, o_t)] \, dh_0 \ldots dh_T,
\label{eq8}
\end{equation}

\noindent and $\mathcal{L}_{q_{\phi}, p_{\psi}}$ denotes as follows:
\begin{equation}
\mathcal{L}_{q_{\phi}, p_{\psi}} = \sum_{t=1}^{T} \mathbb{E}_{q_{\phi}}(h_t | h_{t-1}, o_t) \left[(L_1 + L_2 + L_3)\right],
\label{eq9}
\end{equation}

\noindent where the $L_1$, $L_2$, $L_3$ are respectively defined as:
\begin{equation*}
\begin{aligned}
& L_1 = \log p_\psi(U_t | h_t) \cdot p_\psi(F_t | v_t), \\
& L_2 = \log \left( \frac{p_\psi(E_t | s_t, v_t)}{q_\phi (E_t | s_t, v_t)} \right), \\
& L_3 = \log \left( \frac{p_\psi(h_t | h_{t-1}, P_{t-1})}{q_\phi(h_t | h_{t-1}, o_t)} \right).
\end{aligned}
\end{equation*}

\noindent We parameterize the $p_\psi(E_t | s_t, v_t)$ as $q_\phi (E_t | s_t, v_t)$, making $L_2$ degenerate to 0. At each time stage t, the $L_1$ denotes the reconstruction loss, while the $L_3$ denotes the KL divergence of the hidden variables.

\begin{figure}{}
\centering         
\includegraphics[width=0.45\textwidth]{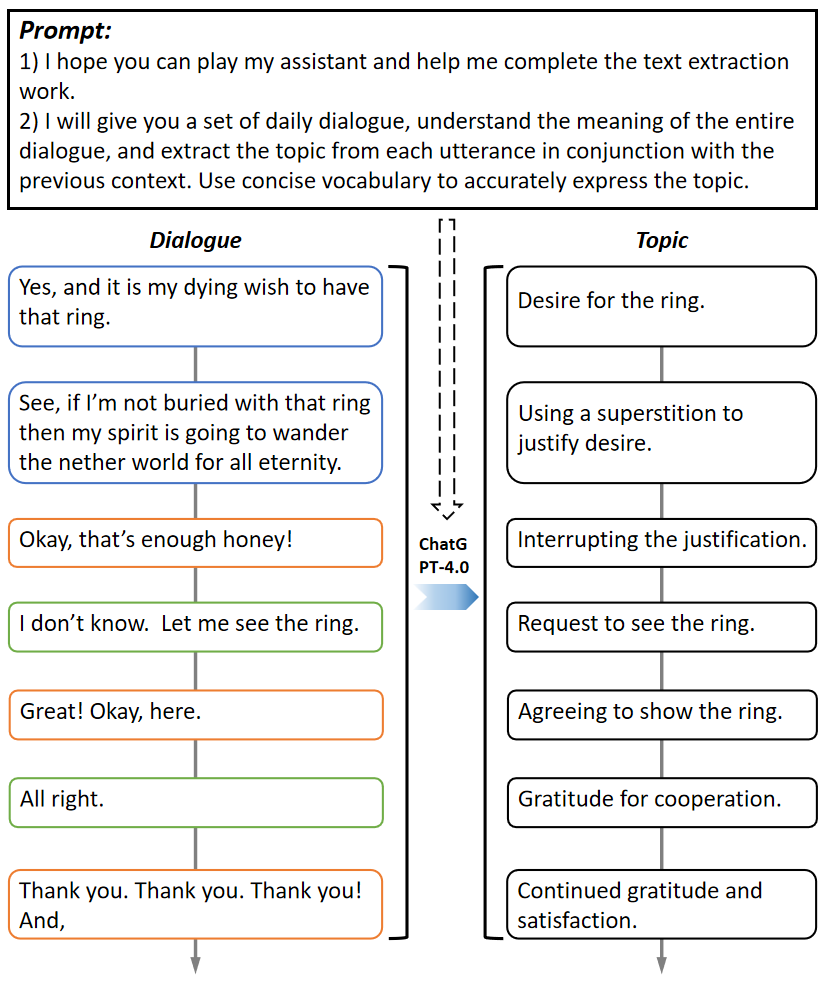}
\caption{The process of extracting conversation topics utilizing ChatGPT-4.0.}
\label{fig4}
\end{figure}

\subsection{\bf Feature extraction based on ChatGPT-4.0}
ChatGPT is a large-scale natural language processing model developed by OpenAI. Following extensive training, ChatGPT exhibits exceptional competence in both comprehending and generating natural language text, making it versatile for diverse natural language processing tasks. For the task of dialogue emotion detection, we rely on ChatGPT for pertinent text feature extraction.

In our approach, obtaining the dialogue utterance's topic as the observation variable is essential. As the conversation unfolds, the topic continuously evolves, necessitating the acquisition of topic features at each time stage. We have selected ChatGPT-4.0 for this task due to its superior contextual understanding in text processing tasks. Specifically, it excels in providing more precise topic information by comprehending preceding dialogues. A visual representation of topic extraction is depicted in Fig.\ref{fig4}.

\section{Experiment}\label{5}
We apply our method to IEMOCAP and MELD datasets. Subsequently, we conducted ablation experiments and robustness analyses. Initially, we present the data conditions and implementation details.

\begin{table}[ht]
\centering
\caption{\textnormal{Statistics of splits in two datasets.}}
\begin{tabular}{>{\centering\arraybackslash}p{1.2cm}|>{\centering\arraybackslash}p{0.5cm}|>{\centering\arraybackslash}p{0.5cm}|>{\centering\arraybackslash}p{0.6cm}|>{\centering\arraybackslash}p{0.5cm}|>{\centering\arraybackslash}p{0.5cm}|>{\centering\arraybackslash}p{0.6cm}|>{\centering\arraybackslash}p{0.8cm}}
\toprule
\multirow{2}{*}{Methods} & \multicolumn{3}{c|}{Dialogue} & \multicolumn{3}{c|}{Utterance} & \multirow{2}{*}{Category} \\
\cmidrule{2-7}
& \multicolumn{2}{c|}{Train \& Val} & Test & \multicolumn{2}{c|}{Train \& Val} &  Test & \\
\midrule
IEMOCAP & \multicolumn{2}{c|}{120} & 31 & \multicolumn{2}{c|}{5810} & 1623 & 6 \\
\rule{0pt}{10pt}MELD & \multicolumn{2}{c|}{1152} & 280 & \multicolumn{2}{c|}{11098} & 2610 & 7 \\
\bottomrule
\end{tabular}
\label{tab1}
\end{table}

\subsection{\bf Datasets}
IEMOCAP \cite{busso2008iemocap} is a multimodal conversation dataset collected by the SAIL laboratory at the University of Southern California. This dataset comprises 151 dialogues, totaling 7433 sentences. The dataset is annotated with six emotion categories: neutral, happy, sad, angry, frustrated, and excited. Non-neutral emotions constitute a majority at 77\%. IEMOCAP stands as the most widely utilized dataset in the field of dialogue emotion detection, characterized by its high-quality annotations. However, it is worth noting that the dataset has a relatively limited data size.

MELD \cite{poria2018meld} is an extension of the EmotionLines dataset \cite{chen2018emotionlines}, augmented with additional audio and video information. The MELD dataset comprises 1433 multi-party dialogues and 13708 utterances extracted from the television series “Friends”. This dataset provides annotations for utterances, categorizing them into seven distinct emotion categories: neutral, joy, surprise, sadness, anger, disgust, and fear. Non-neutral emotions constitute a majority at 53\%. MELD is distinguished by its high-quality annotations and the incorporation of multimodal data. Statistics of splits in two datasets are shown in Table.\ref{tab1}.

\subsection{\bf Baseline and State-of-the-Arts}
BC-LSTM \cite{poria2017context} proposes a bidirectional LSTM network to capture contextual content, generate context aware discourse representations, and utilize them for sentiment classification. However, this model only considers context dependencies and ignores speaker dependencies.

CMN \cite{hazarika2018conversational} utilizes the dependency relationships of speakers to identify emotions. It designs different GRU networks for two speakers to model utterance context from dialogue history and inputs current utterances into different memory networks to learn utterance representation. However, this model is limited to conversations with only two participants.

ICON \cite{hazarika2018icon} is an extension of CMN that utilizes another GRU to connect the GRU network outputs of two speakers for explicit speaker modeling. Similarly, it can only model conversations between two speakers.

ConGCN \cite{zhang2019modeling} constructs a heterogeneous graph convolutional neural network. The utterances and speakers of a conversation are represented by nodes. The graph contains edges between utterance nodes and edges between utterance and speaker nodes. This method models context and speaker-sensitive dependencies for emotion detection.

KET \cite{zhong2019knowledge} is a knowledge-enriched transformer for emotion detection. It dynamically leverages common sense knowledge by using a context aware graph attention mechanism.

DialogueRNN \cite{majumder2019dialoguernn} is a benchmark model for dialogue emotion detection. It is a recurrent network that utilizes three GRU networks to model the speaker, the context from the preceding utterances, and the emotion of the preceding utterance to track the emotion of each speaker.

DialogueGCN \cite{ghosal2019dialoguegcn} represents the structure and interaction of a conversation by constructing a graph structure between participants and utilizes graph convolutional neural networks for message passing and feature learning of the graph structure.

KES \cite{ren2021utilizing} utilizes a self-attention layer specialized for enhanced semantic text features with external commonsense knowledge and two networks based on LSTM for tracking individual internal state and context external state to learn interactions between interlocutors participating in a conversation.

TL-ERC \cite{hazarika2021conversational} utilizes a generative conversational model to transfer emotional knowledge. It transfers the parameters of a trained hierarchical model to an emotion detection classifier.

SS-HiT \cite{zhang2022sentiment} is a novel semantic and sentiment hierarchical transformer for emotion detection. Each utterance token is represented as matrices with both semantic and sentiment word embeddings. Then, fuse utterance tokens as token features to further capture the long dependent utterance-level information through transformer encoders for prediction.

Sentic GAT \cite{tu2022context} is a context- and sentiment-aware framework, which designs a dialogue transformer (DT) network with hierarchical multihead attention to capture the intra- and inter-dependency relationship in context. In addition, common-sense knowledge is dynamically represented by the context- and sentiment-aware graph attention mechanism based on sentimental consistency.

EmpaGen \cite{wang2022multiturn} proposes an auxiliary empathic multiturn dialogue generation to enhance dialogue emotion understanding, which is the first attempt to utilize empathy-based dialogue generation for the emotion detection task.

GGCN \cite{nie2023long} is a growing graph convolution network that proposes a prior knowledge extraction framework to get auxiliary information. The method considers utterance connections and emotion accumulation for emotional detection.

IEIN \cite{lu2020iterative} is an iterative emotional interaction model that utilizes iteratively predicted emotional labels instead of real emotional labels to continuously correct predictions and provide feedback on inputs during the iteration process.

SDTN \cite{chen2023sdtn} dynamically tracks the local and global speaker states as the conversation progresses to capture implicit stimulation of emotional shift. It mainly contains the speaker interaction tracker based on GRU and the emotion state decoder module based on the conditional random field.

\begin{table*}[ht]
\centering
\caption{\textnormal{Comparison with state-of-the-art methods on the IEMOCAP dataset.}}
\begin{tabular}{>{\centering\arraybackslash}p{2.3cm}|>{\centering\arraybackslash}p{0.5cm} >{\centering\arraybackslash}p{0.5cm}|>{\centering\arraybackslash}p{0.5cm} >{\centering\arraybackslash}p{0.5cm}|>{\centering\arraybackslash}p{0.5cm} >{\centering\arraybackslash}p{0.5cm}|>{\centering\arraybackslash}p{0.5cm} >{\centering\arraybackslash}p{0.5cm}|>{\centering\arraybackslash}p{0.5cm} >{\centering\arraybackslash}p{0.5cm}|>{\centering\arraybackslash}p{0.5cm} >{\centering\arraybackslash}p{0.5cm}|>{\centering\arraybackslash}p{0.5cm} >{\centering\arraybackslash}p{0.5cm}}
\toprule
\multirow{2}{*}{Methods} & \multicolumn{2}{c|}{Happy} & \multicolumn{2}{c|}{Sad} & \multicolumn{2}{c|}{Neutral} & \multicolumn{2}{c|}{Angry} & \multicolumn{2}{c|}{Excited} & \multicolumn{2}{c|}{Frustrated} & \multicolumn{2}{c}{\textbf{\textit{Avg.}}} \\
\cmidrule{2-15}
& Acc. & F1 & Acc. & F1 & Acc. & F1 & Acc. & F1 & Acc. & F1 & Acc. & F1 & Acc. & F1 \\
\midrule
BC-LSTM \cite{poria2017context} & 18.8 & 28.7 & 67.8 & 71.2 & 60.4 & 54.0 & 59.4 & 61.4 & 57.9 & 63.0 & 68.2 & 60.9 & 59.1 & 58.4 \\
\rule{0pt}{10pt}CMN \cite{hazarika2018conversational} & 25.0 & 30.4 & 55.9 & 62.4 & 52.9 & 52.4 & 61.8 & 59.8 & 55.5 & 60.3 & \textbf{71.1} & 60.7 & 56.6 & 56.1 \\
\rule{0pt}{10pt}ICON \cite{hazarika2018icon} & 22.2 & 29.9 & 58.8 & 64.6 & 62.8 & 57.4 & 64.7 & 63.0 & 58.9 & 63.4 & 67.2 & 60.8 & 59.1 & 58.5 \\
\rule{0pt}{10pt}KET \cite{zhong2019knowledge} & - & - & - & - & - & - & - & - & - & - & - & - & - & 59.6 \\
\rule{0pt}{10pt}DialogueRNN \cite{majumder2019dialoguernn} & 25.7 & 33.2 & 75.1 & 78.8 & 58.6 & 59.2 & 64.7 & \textbf{65.3} & \textbf{80.3} & 71.9 & 61.2 & 58.9 & 63.4 & 62.8 \\
\rule{0pt}{10pt}DialogueGCN \cite{ghosal2019dialoguegcn} & 40.6 & 42.8 & \textbf{89.1} & 84.5 & 61.9 & 63.5 & \textbf{67.5} & 64.2 & 65.5 & 63.1 & 64.2 & 67.0 & 65.3 & 64.2 \\
\rule{0pt}{10pt}KES \cite{ren2021utilizing} & - & 47.7 & - & 84.6 & - & 64.3 & - & 62.5 & - & 73.3 & - & 63.5 & - & 66.3 \\
\rule{0pt}{10pt}IEIN \cite{lu2020iterative} & - & 53.2 & - & 77.2 & - & 61.3 & - & 61.5 & - & 69.2 & - & 60.9 & - & 64.4 \\
\rule{0pt}{10pt}GGCN \cite{nie2023long} & \textbf{56.6} & 49.8 & 85.3 & \textbf{85.5} & 62.6 & 63.8 & 61.3 & 63.1 & 73.5 & 75.2 & 65.0 & 67.4 & 67.9 & 68.7 \\
\rule{0pt}{10pt}SS-HiT \cite{zhang2022sentiment} & - & - & - & - & - & - & - & - & - & - & - & - & - & 61.0 \\
\rule{0pt}{10pt}Our & 53.5 & \textbf{54.4} & 78.8 & 78.2 & \textbf{63.5} & \textbf{65.1} & 66.0 & 64.4 & 76.3 & \textbf{75.7} & 69.8 & \textbf{69.3} & \textbf{68.8} & \textbf{68.9} \\
\bottomrule
\end{tabular}
\label{tab2}
\end{table*}

\subsection{\bf Implementation Details}
In this section, we will furnish a comprehensive overview of the implementation of our proposed method. We pre-extract topic features and personal attribute features from the prior dialogue content. For the extraction of personal attribute features, we design an LSTM network and specify the feature vector dimension as 64.

We utilize ChatGPT-4.0 API with Python 3 on a single Nvidia RTX 3060 GPU and an Intel i7 CPU. We configure ChatGPT-4.0 as a text work assistant for the task of text extraction. Following the role assignment, we provide the prompt: “I will give you a set of daily dialogue, understand the meaning of the entire dialogue, and extract the topic from each utterance in conjunction with the previous context. Use concise vocabulary to accurately express the topic.” Additionally, we specify detailed format requirements to facilitate future text collection and dataset construction. To ensure a comprehensive understanding of dialogue content and context, we input entire dialogues into the model, instructing it to extract topic information for each utterance. However, in the case of lengthy dialogues with numerous utterances, the model occasionally miscounted them, leading to a mismatch between the number of generated topics and utterances. To address this, we segment long dialogues into batches, each containing 20 utterances. This approach allows ChatGPT-4.0 to better grasp the context and accurately determine the number of utterances. We apply the bert-base-uncased model for processing English text to convert topic text information into 768-dimension feature vectors for subsequent experiments, which is one of the pre-trained BERT models \cite{devlin2018bert}. Then we utilize two linear layers to reduce the dimensionality of the feature vectors to 64 dimensions.

Then, we construct the Dynamic Causal Disentanglement Model by combining all observed variables. The parameter matrices of the observation variable extraction model and the model are randomly initialized and continuously optimized during training. Throughout the training process, we utilize Adam as the optimizer with the weight attenuation set to 0.00005 and the learning rate set to 0.001. The model is trained for 80 epochs. We develop the evaluations with Python 3 and PyTorch 1.13.0 on a single Nvidia RTX 3060 GPU and an Intel i7 CPU.

\subsection{\bf Comparison with State-of-the-Arts}
We compare the performance of our proposed model with the current state-of-the-art methods. The experimental results show that our model performs better than other state-of-the-art models on two benchmark datasets. 

\textit{IEMOCAP:} 
IEMOCAP is a dyadic interactive dialogues dataset consisting of long conversations and it contains many non-neutral emotions. We evaluate our model on the IEMOCAP dataset, and the experiment results are shown in Table.\ref{tab2}. 
Our model achieves a new state-of-the-art F1-score of 68.9\% and accuracy of 68.8\% for the emotion detection task on the IEMOCAP dataset. We compare the results of our proposed model with the state-of-the-art methods. Although our model still needs to improve the performance in some emotional categories, it is worth noting that our model has made significant improvements in both \textit{Happy} and \textit{Frustrated} emotions. 
Compared with the IEIN method, our model demonstrates a critical improvement of 1.2\% in the average F1-score of \textit{Happy} emotion. 
Compared with the state-of-the-art methods, our model achieves improvements in the accuracy of 0.7\% and the average F1-score of 0.8\% for the \textit{Neutral} emotion. 
Compared with the GGCN method, our model achieves a significant improvement of 0.5\% and 1.9\% in the average F1-score of the \textit{Excited} and \textit{Frustrated} emotions.
In general, our approach outperforms the state-of-the-art methods.

\textit{MELD:} 
The MELD dataset comprises multiparty conversations extracted from television series. Notably, many MELD dialogues involve more than five participants, with a maximum of nine, resulting in limited utterances per participant and posing challenges for contextual dependency modeling. Furthermore, the dataset exhibits a substantial presence of non-neutral emotions. In contrast to the IEMOCAP dataset, MELD dialogues are notably shorter, with an average length of 10 utterances, intensifying the challenges in emotion detection. Importantly, we find that the CMN and ICON methods, designed for dyadic conversations, are not suitable for MELD. These emotion detection models achieve poor results on this dataset.

Our model achieves a new state-of-the-art F1-score of 67.5\%, and our model obtains the best results among the compared methods, the experiment results are shown in Table.\ref{tab3}. Our model has made significant progress in \textit{Anger}, \textit{Joy}, and \textit{Surprise} emotions. 
Although the F1-score of our model is not as good as IEIN and GGCN, compared to most methods, we still make progress in \textit{Disgust} and \textit{Fear} emotions.
We observe that the topic features extracted by ChatGPT-4.0 substantially enhance our model's performance, enabling a deeper understanding of utterance meanings and the speakers' psychology in brief conversations involving multiple participants. Consequently, our model demonstrates exceptional performance on the MELD dataset.

\begin{table*}[ht]
\centering
\caption{\textnormal{Comparison with state-of-the-art methods on the MELD dataset.}}
\begin{tabular}{>{\centering\arraybackslash}p{2.3cm}|>{\centering\arraybackslash}p{1cm}|>{\centering\arraybackslash}p{1cm}|>{\centering\arraybackslash}p{1cm}|>{\centering\arraybackslash}p{1cm}|>{\centering\arraybackslash}p{1cm}|>{\centering\arraybackslash}p{1cm}|>{\centering\arraybackslash}p{1cm}|>{\centering\arraybackslash}p{1cm}}
\toprule
Methods & Anger & Disgust & Fear & Joy & Neutral & Sadness & Surprise & \textbf{\textit{Avg.}} \\
\midrule
BC-LSTM \cite{poria2017context} & 40.5 & 0.0 & 0.0 & 52.9 & 76.0 & 24.2 & 47.6 & 57.1 \\
\rule{0pt}{10pt} ConGCN \cite{zhang2019modeling} & 46.8 & 10.6 & 8.7 & 53.1 & 76.7 & 28.5 & 50.3 & 59.4 \\
\rule{0pt}{10pt} KET \cite{zhong2019knowledge} & - & - & - & - & - & - & - & 58.2 \\
\rule{0pt}{10pt} DialogueRNN \cite{majumder2019dialoguernn} & 46.8 & 0.81 & 0.0 & 54.6 & 76.2 & 26.3 & 49.6 & 58.7 \\
\rule{0pt}{10pt} DialogueGCN \cite{ghosal2019dialoguegcn} & 43.0 & 1.2 & 1.0 & 53.6 & 76.0 & 24.3 & 46.4 & 57.5\\
\rule{0pt}{10pt} KES \cite{ren2021utilizing} & - & - & - & - & - & - & - & 66.5\\
\rule{0pt}{10pt} TL-ERC \cite{hazarika2021conversational} & 48.9 & 8.7 & 11.2 & 60.4 & 80.4 & 25.6 & 53.8 & 61.9 \\
\rule{0pt}{10pt} SS-HiT \cite{zhang2022sentiment} & - & - & - & - & - & - & - & 59.3 \\
\rule{0pt}{10pt} Sentic GAT \cite{tu2022context} & - & - & - & - & - & - & - & 59.2 \\
\rule{0pt}{10pt} EmpaGen \cite{wang2022multiturn} & 50.2 & 10.9 & 9.5 & 61.8 & 79.2 & 27.6 & 55.6 & 62.4 \\
\rule{0pt}{10pt} IEIN \cite{lu2020iterative} & 48.9 & \textbf{19.4} & 3.3 & 56.6 & 77.5 & 23.6 & 53.7 & 60.7 \\
\rule{0pt}{10pt} GGCN \cite{nie2023long} & 49.8 & 15.5 & \textbf{14.3} & 63.1 & \textbf{82.9} & \textbf{41.2} & 66.3 & 67.3\\
\rule{0pt}{10pt} SDTN \cite{chen2023sdtn} & - & - & - & - & - & - & - & 66.1\\
\rule{0pt}{10pt}Our & \textbf{50.9} & 16.2 & 10.8 & \textbf{65.0} & 82.6 & 38.6 & \textbf{67.4} & \textbf{67.5} \\
\bottomrule
\end{tabular}
\label{tab3}
\end{table*}

\begin{table*}[t]
\centering
\caption{\textnormal{Experimental results of ablation study.}}
\begin{tabular}{>{\centering\arraybackslash}p{2.4cm}|>{\centering\arraybackslash}p{2cm}|>{\centering\arraybackslash}p{2.4cm}|>{\centering\arraybackslash}p{2cm}|>{\centering\arraybackslash}p{1.4cm}|>{\centering\arraybackslash}p{1.4cm}|>{\centering\arraybackslash}p{1.4cm}}
\toprule
\multicolumn{4}{c|}{Components} & \multicolumn{2}{c|}{IEMOCAP} & MELD \\
\midrule
Topic(ChatGPT-4.0) & Topic(LSTM) & Personal Attribute & Disentanglement & Acc. & W-Avg F1 & W-Avg F1 \\
\midrule
\Checkmark & \XSolidBrush & \Checkmark & \XSolidBrush & 59.4 & 59.7 & 59.4 \\
\rule{0pt}{10pt}\XSolidBrush & \XSolidBrush & \XSolidBrush & \Checkmark & 62.8 & 62.1 & 62.2 \\
\rule{0pt}{10pt}\XSolidBrush & \Checkmark & \XSolidBrush & \Checkmark & 64.0 & 64.1 & 63.7 \\
\rule{0pt}{10pt}\XSolidBrush & \XSolidBrush & \Checkmark & \Checkmark & 65.7 & 65.8 & 65.0 \\
\rule{0pt}{10pt}\XSolidBrush & \Checkmark & \Checkmark & \Checkmark & 67.8 & 68.0 & 66.7 \\
\rule{0pt}{10pt}\Checkmark & \XSolidBrush & \Checkmark & \Checkmark & \textbf{68.8} & \textbf{68.9} & \textbf{67.5} \\
\bottomrule
\end{tabular}
\label{tab4}
\end{table*}

\begin{figure}[ht]      
\centering
  \subfigure[The performance of different methods on IEMOCAP dataset]{
  \includegraphics[width=0.4\linewidth]{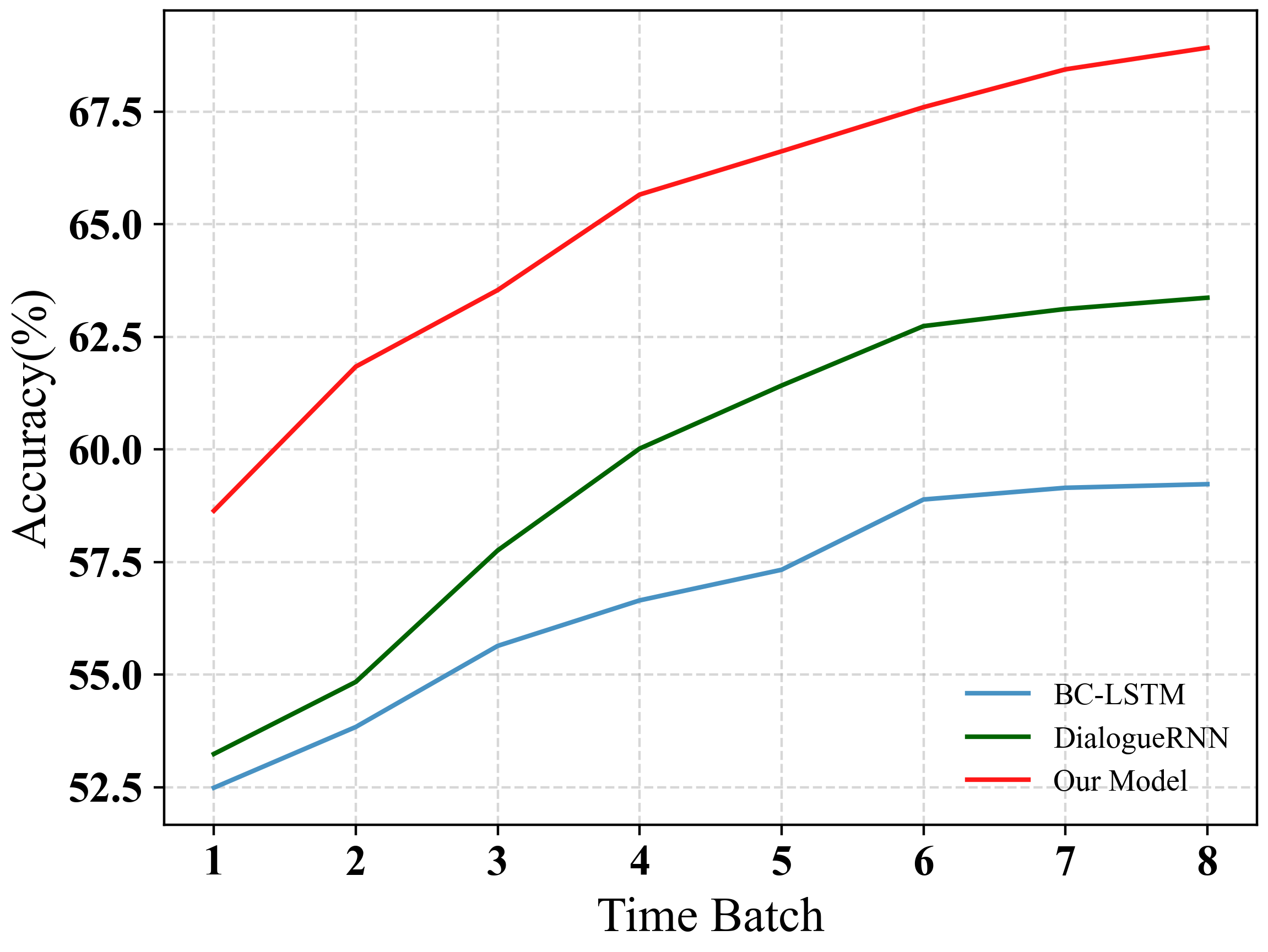}
  \hspace{0.05\linewidth} 
}
    \subfigure[The performance of different methods on MELD dataset]{
  \includegraphics[width=0.4\linewidth]{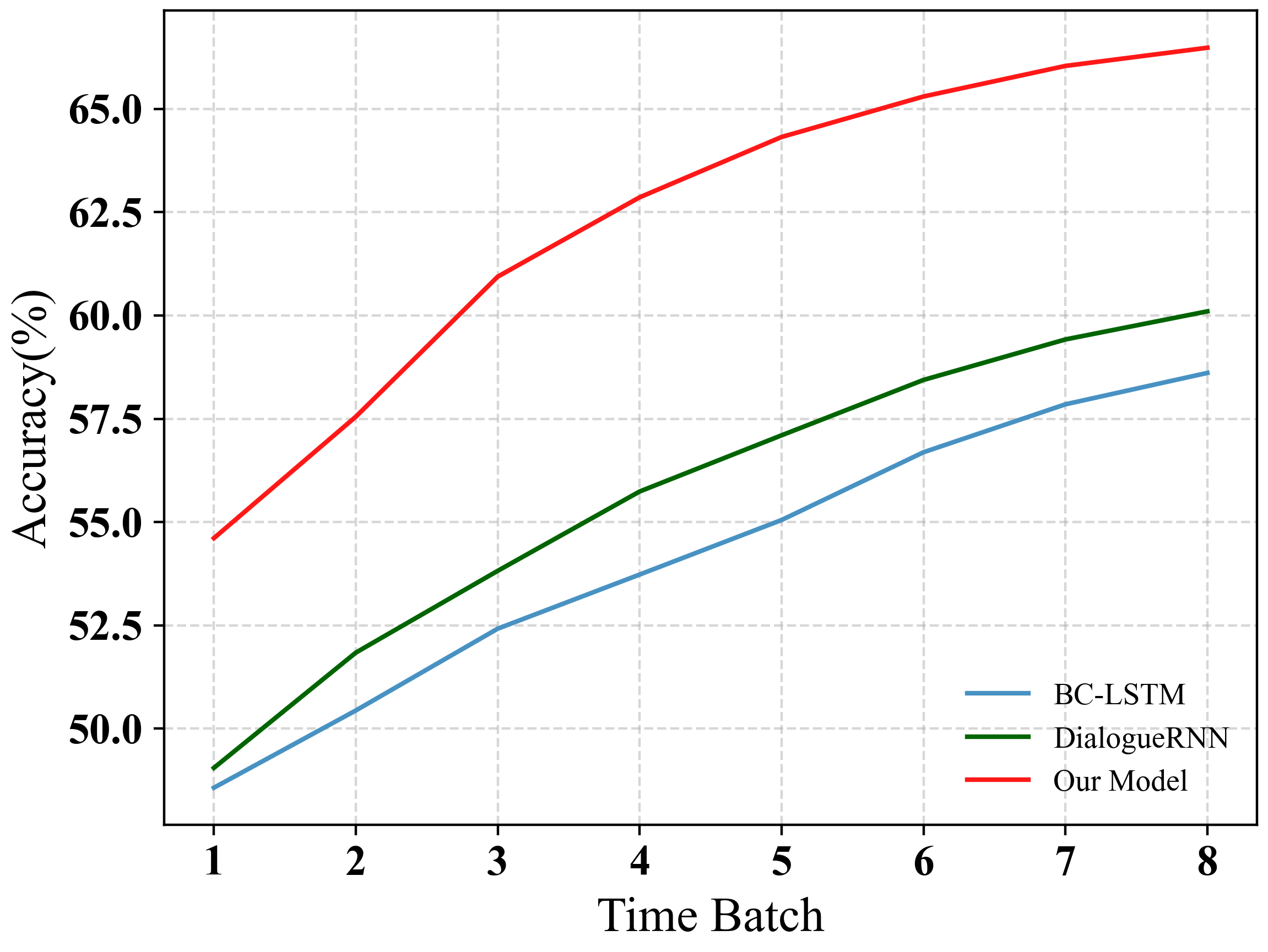}
  \hspace{0.05\linewidth} 
}
\caption{Prediction accuracy of long dialogue utterances on the IEMOCAP and MELD datasets. The blue, green, and red lines represent BC-LSTM, DialogueRNN, and our model respectively.}
\label{fig5}
\end{figure}

\subsection{\bf Ablation Study}
To underscore the effectiveness of hidden variables disentanglement and the significance of external variables in our approach, we undertake an ablation study.

The comparison results are in Table.\ref{tab4}, we conclude that leveraging relevant hidden variables derived from disentanglement aids in emotion detection. Our approach demonstrates significant improvements in accuracy and F1-score when compared to models lacking disentanglement. The result validates our idea and also proves the effectiveness of disentanglement. 

We observe that when employing LSTM-extracted topic features and personal attribute features separately, the models incorporating personal attribute features outperformed those relying on topic features by 1.7\% and 1.3\% in terms of F1-scores for both datasets, respectively.
Our analysis suggests that personal attribute features, serving as auxiliary variables in the propagation of hidden variables, elucidate the distribution of hidden variables, thereby exerting a more pronounced influence on result enhancement. We find that in comparison to the model using theme features extracted by LSTM, the models incorporating theme features extracted by ChatGPT-4.0 and BERT exhibited improvements of 0.9\% and 0.8\% in F1-score on both datasets, respectively, demonstrating the superior ability of ChatGPT-4.0 in text processing.

The model combining two observation variables and disentanglement achieves the best results, which verifies the effectiveness of our proposed method and all observed variables. Overall, the introduction of observed variables and the disentanglement of hidden variables collectively enhance the accuracy of emotion detection.

\begin{figure*}[ht]
\centering
  \subfigure[Pre-processing data distribution]{
  \begin{minipage}{0.31\linewidth}
  \centering
  \includegraphics[width=1.1\linewidth]{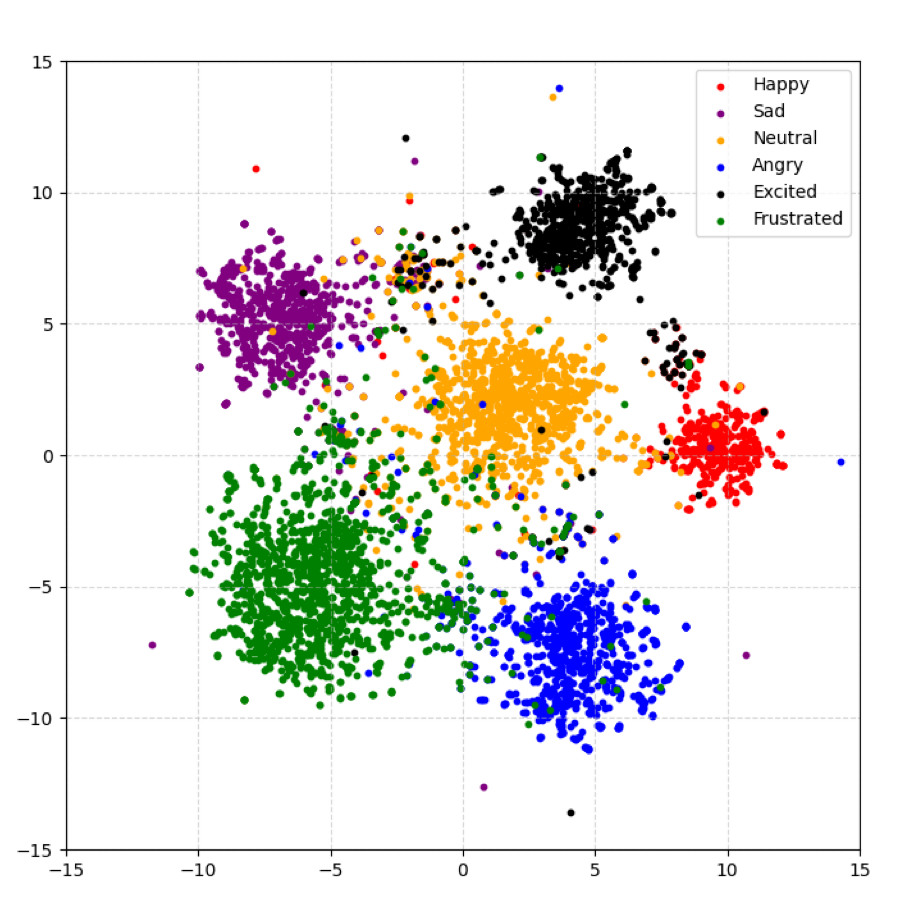}
  \end{minipage}
}
    \subfigure[Data distribution learned by $s$, $v$, and $z$]{
  \begin{minipage}{0.31\linewidth}
  \centering
  \includegraphics[width=1.1\linewidth]{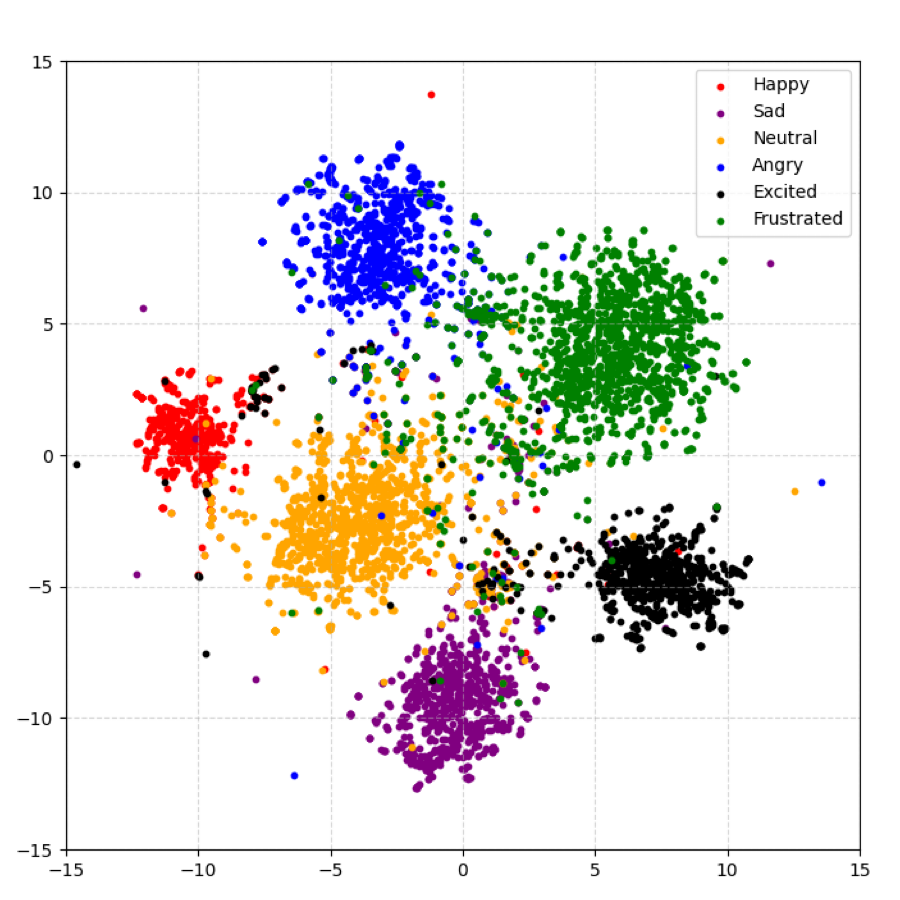}
  \end{minipage}
}
    \subfigure[Data distribution learned by $s$ and $v$]{
  \begin{minipage}{0.31\linewidth}
  \centering
  \includegraphics[width=1.1\linewidth]{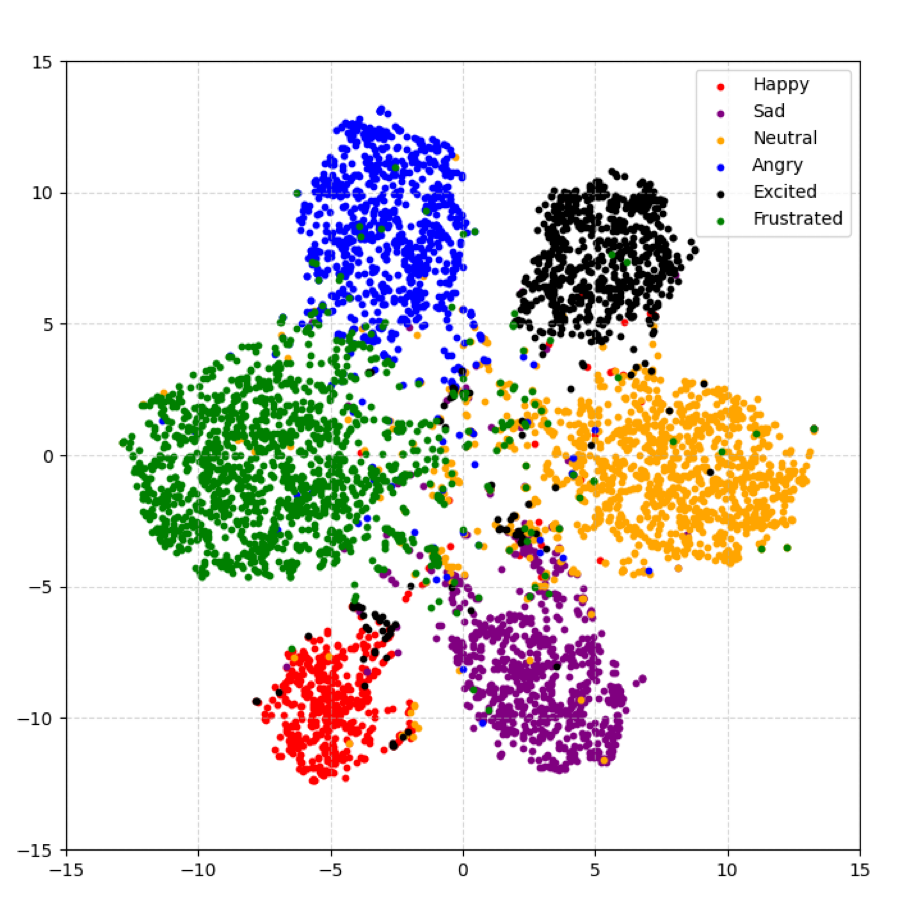}
  \end{minipage}
  }
\caption{The feature mappings of high-dimensional vectors obtained by t-SNE. Dots of various colors represent utterances with distinct emotion labels.}
\label{fig6}
\end{figure*}

\subsection{\bf Discussion on Prediction Accuracy of Long Dialogue Utterances}
To further assess the stability of our proposed model, we calculate the accuracy of emotion detection within each period, as illustrated in Fig.\ref{fig5}. Specifically, for the IEMOCAP dataset, we divide every five utterances into a time batch, and due to the extended length of dialogues in this dataset, we analyze the first forty utterances. In the case of MELD, a dataset featuring shorter dialogues, we segment each utterance into a time batch and analyze the first eight utterances.

We observe that the emotion detection methods BC-LSTM and DialogueRNN exhibit lower accuracy in the early stages of a dialogue, with accuracy gradually improving as the dialogue progresses. The results suggest that these contextual modeling methods require a greater number of utterances to comprehend ongoing dialogues due to the changing dialogue topics and contexts. In contrast, our method effectively mitigates this challenge. By disentangling hidden variables related to emotions at the outset of the dialogue, our model achieves high accuracy in the early time batch. Furthermore, our prediction accuracy remains stable and consistently high, facilitated by the incorporation of dialogue topics and personal attributes.

\begin{table}[t]
\centering
\caption{\textnormal{Prediction results of different hidden variables.}}
\begin{tabular}{>{\centering\arraybackslash}p{1cm}|>{\centering\arraybackslash}p{1cm}|>{\centering\arraybackslash}p{1cm}|>{\centering\arraybackslash}p{1cm}|>{\centering\arraybackslash}p{1cm}|>{\centering\arraybackslash}p{1cm}}
\toprule
\multicolumn{3}{c|}{Variables} & \multicolumn{2}{c|}{IEMOCAP} & MELD \\
\midrule
$s$ & $v$ & $z$ & Acc. & F1 &  F1 \\
\midrule
\XSolidBrush & \XSolidBrush & \Checkmark & 59.8 & 60.2 & 59.8 \\
\rule{0pt}{10pt} \XSolidBrush & \Checkmark & \XSolidBrush & 66.1 & 66.4 & 66.4 \\
\rule{0pt}{10pt} \Checkmark & \XSolidBrush & \XSolidBrush & 66.4 & 66.5 & 66.8 \\
\rule{0pt}{10pt} \Checkmark & \XSolidBrush & \Checkmark & 62.9 & 63.2 & 63.4 \\
\rule{0pt}{10pt} \XSolidBrush & \Checkmark & \Checkmark & 63.8 & 63.8 & 63.7 \\
\rule{0pt}{10pt} \Checkmark & \Checkmark & \XSolidBrush & \textbf{68.8} & \textbf{68.9} & \textbf{67.5} \\
\bottomrule
\end{tabular}
\label{tab5}
\end{table}

\begin{figure*}[ht]
\centering
  \subfigure[BC-LSTM model on IEMOCAP]{
  \begin{minipage}{0.31\linewidth}
  \centering
  \includegraphics[width=1.1\linewidth]{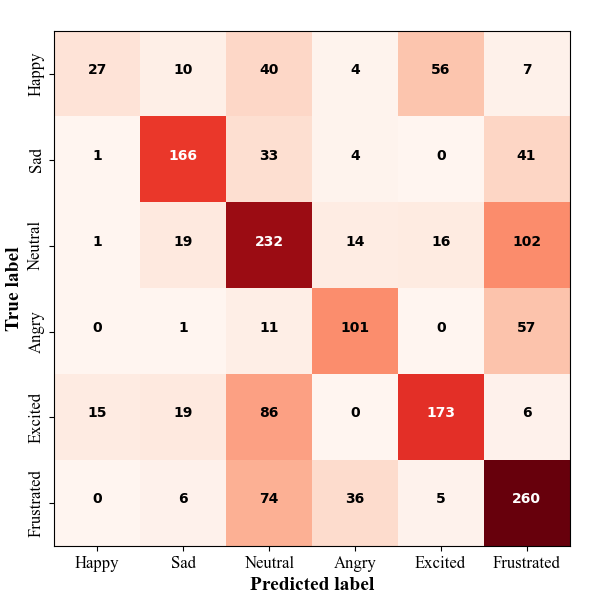}
  \end{minipage}
}
    \subfigure[DialogueRNN model on IEMOCAP]{
  \begin{minipage}{0.31\linewidth}
  \centering
  \includegraphics[width=1.1\linewidth]{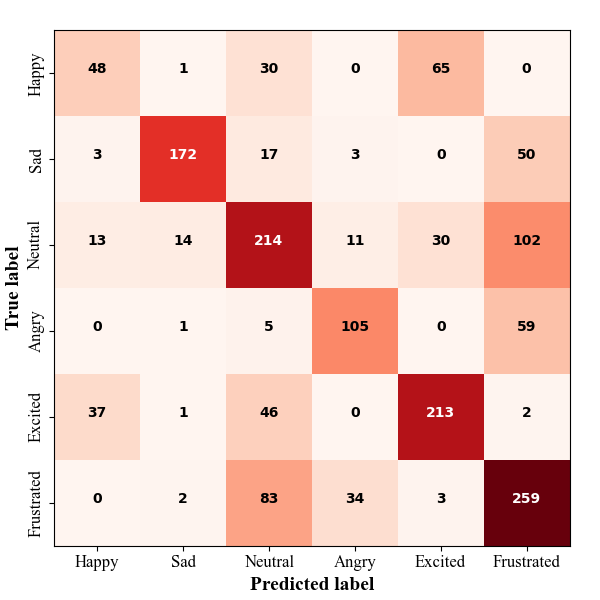}
  \end{minipage}
}
    \subfigure[Our model on IEMOCAP]{
  \begin{minipage}{0.31\linewidth}
  \centering
  \includegraphics[width=1.1\linewidth]{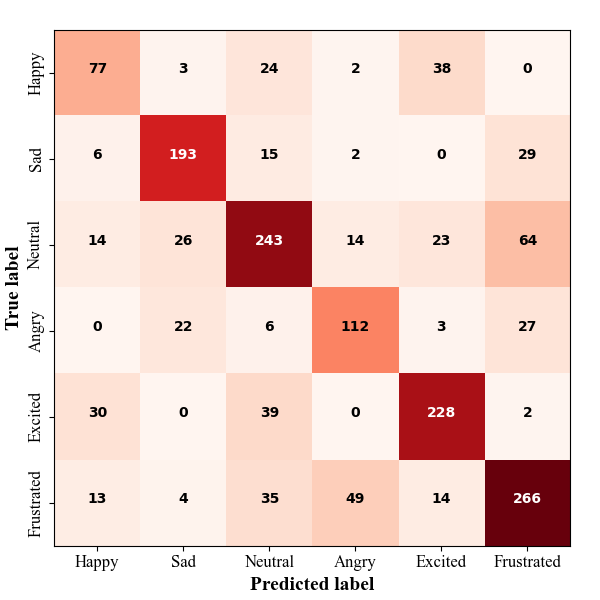}
  \end{minipage}
  }
    \subfigure[BC-LSTM model on MELD]{
  \begin{minipage}{0.31\linewidth}
  \centering
  \includegraphics[width=1.1\linewidth]{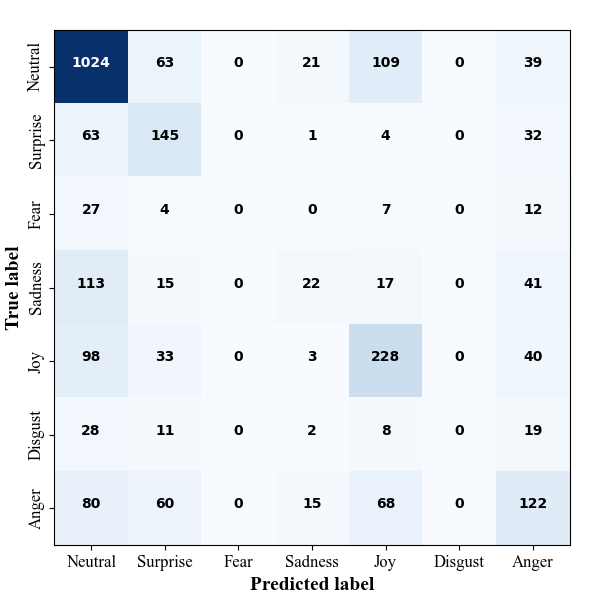}
  \end{minipage}
}
    \subfigure[DialogueRNN model on MELD]{
  \begin{minipage}{0.31\linewidth}
  \centering
  \includegraphics[width=1.1\linewidth]{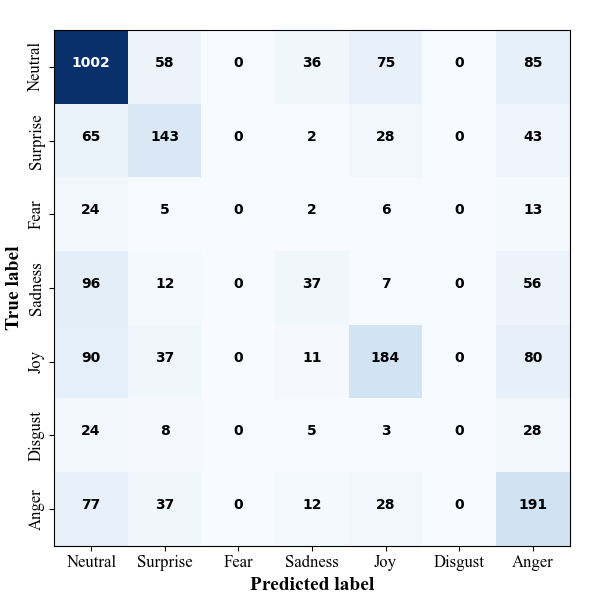}
  \end{minipage}
}
    \subfigure[Our model on MELD]{
  \begin{minipage}{0.31\linewidth}
  \centering
  \includegraphics[width=1.1\linewidth]{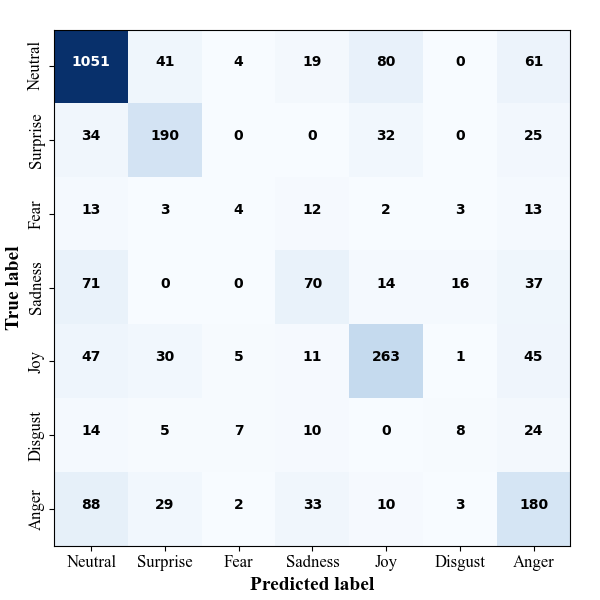}
  \end{minipage}
  }
\caption{Confusion matrix graph of emotion detection with different methods on the IEMOCAP and MELD datasets.}
\label{fig7}
\end{figure*}

\subsection{\bf Effectiveness of Disentanglement}
To demonstrate the reliability of the related variables after disentanglement, we implement a further validation method for $s$, $v$, and $z$. 
Specifically, we first train the Dynamic Causal Disentanglement Model and the observation variable extraction model to obtain the distribution of hidden variables. 
Then, we independently train six independent classifiers for prediction by different combinations of hidden variables, and the experiment results are shown in Table.\ref{tab5}. 
We find that the results of the prediction containing $z$ are lower than those without $z$, which proves the existence of spurious correlation information in the utterances.
The experiments conducted with $s$ and $v$ both yield favorable results.
The accuracy and F1 score obtained from variable $s$ surpass those extracted from variable $v$ on two datasets, suggesting that $s$ harbors more salient feature information about emotional factors.
The experimental results obtained by combining $s$ and $v$ are the best, affirming the validity and effectiveness of our approach.

To intuitively observe the influence of hidden variables on emotional features, we visualize emotional features acquired under varying conditions on the IEMOCAP dataset to examine emotion distribution within the feature space.
Employing the t-SNE method \cite{van2008visualizing}, we project high-dimensional text features into a two-dimensional space, utilizing distinct colors for emotion labels, as depicted in Fig.\ref{fig6}.
In the two-dimensional space after dimensionality reduction, the inter-point distances serve to reflect high-dimensional data similarities.
Observing the distribution of the text feature vectors obtained by pre-processing, we find most utterances with different emotions are bounded by each other, but a small number of utterances still exhibit a messy distribution.
Simultaneously, we observe that utterances with non-neutral emotions and utterances with neutral emotions are close in two-dimensional space, implying their higher similarity in high-dimensional feature vectors.
In contrast, the performance of the text feature vectors learned by the hidden variable $s$, $v$, and $z$ demonstrates some enhancement. Utterances with neutral emotions tend to stay away from utterances with other emotions. However, the introduction of unrelated variable $z$ constrains more precise boundary distinctions.
Notably, we find that the text feature vectors learned by relevant variables $s$ and $v$ with the separation of the unrelated variable $z$ achieve the best performance. The distinction between utterances with different emotions is more accurate, and the distance in the two-dimensional space is far, indicating the low similarity of high-dimensional features. The result further validates our perspective.

\begin{figure*}[ht]
\centering         
\includegraphics[width=0.8\linewidth]{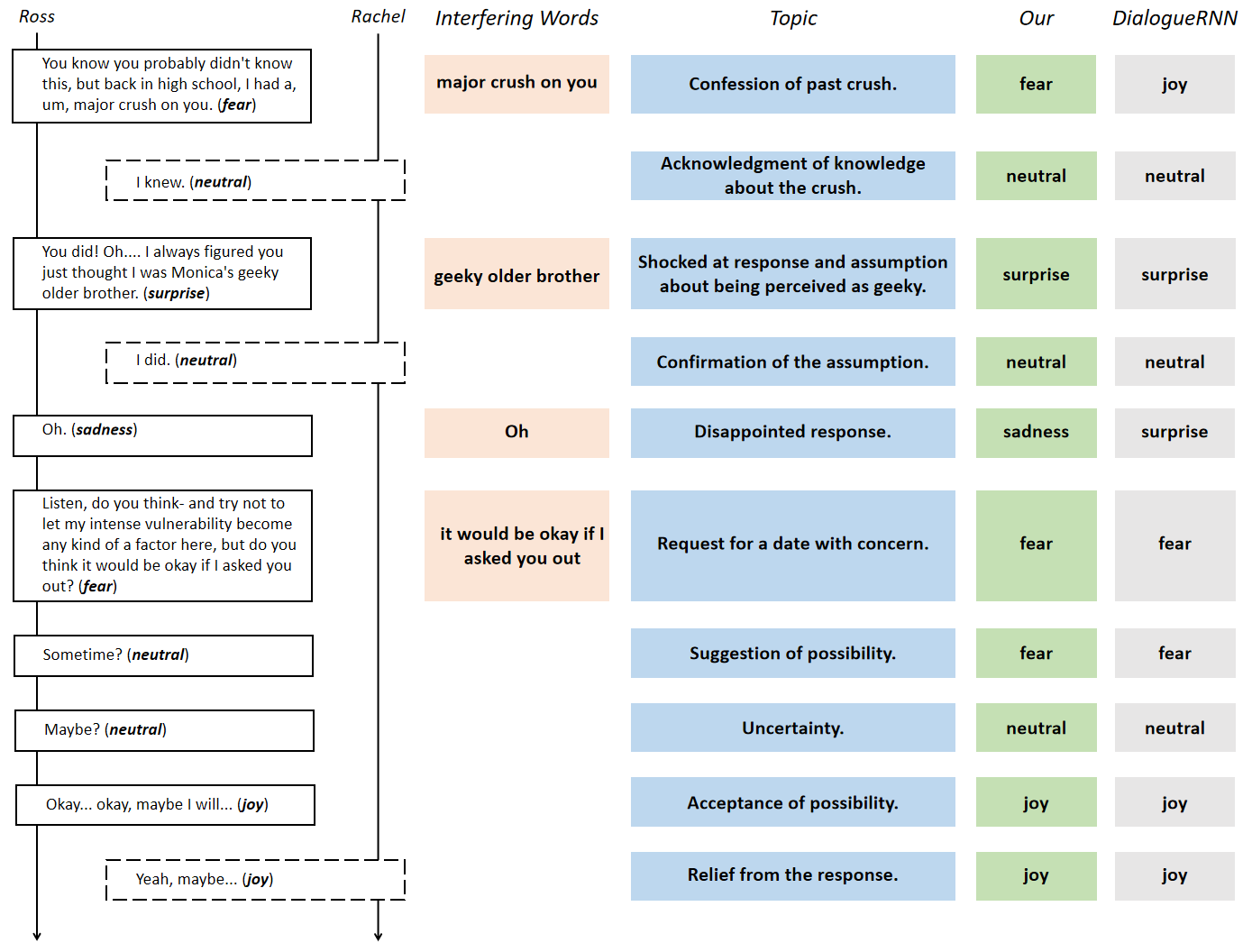}
\caption{The importance of unrelated word separation and topic extraction in dialogue emotion detection. The phrase “crush on you” is a mention of the past rather than an expression of love. The reply “oh” is a response of sadness rather than a common surprise. Our method can comprehensively understand the meaning of dialogue utterances.}
\label{fig8}
\end{figure*}

\subsection{\bf Visualization of Confusion Matrix and Analysis}
We show the visualization to observe the comparison of emotion classification results between different methods. The confusion matrixes of different methods on the IEMOCAP and MELD datasets are shown in Fig.\ref{fig7}. 

In the case of the IEMOCAP dataset, experimental results reveal a notably high misprediction rate between the \textit{Neutral} and \textit{Frustrated} emotions. Our analysis indicates that, in some instances, humans may not express their frustration directly in their utterances, making it significantly different from sadness. This observation underscores that a speaker's emotion cannot be fully comprehended based solely on dialogue utterances.
Our method makes up for the shortcomings of BC-LSTM, DialogueRNN, and other methods and we achieve a significant improvement of 0.8\% and 1.9\% in the average F1-score of the \textit{Neutral} and \textit{Frustrated} emotion, which demonstrates our model has a more comprehensive understanding of dialogue utterance. Furthermore, our method enhances F1-score accuracy across other emotional categories. We attribute this improvement to the separation of interfering words and the utilization of ChatGPT-4.0 for topic extraction.
In the MELD dataset, the imbalance in emotion labels poses a challenge to emotion detection, leading to many non-neutral emotions being incorrectly classified as \textit{Neutral} in the validation set. Through a thorough examination of the confusion matrix, We observe that our model substantially enhances the accuracy of \textit{Fear} and \textit{Disgust} emotions detecting, in contrast to other models incapable of detecting these specific emotions. Our model achieves remarkable enhancements in accuracy and F1-scores across multiple emotional categories.

We choose a dialogue from the MELD dataset to validate our viewpoint, as shown in Fig.\ref{fig8}.
The phrase “crush on you” mentioned in the first sentence does not convey an expression of love but rather refers to a concerning reference to events from the past, which may mislead our prediction. 
Within the dialogue, when Ross responds with “oh” upon being characterized as geeky, this response initially appears surprising. However, considering the context from the preceding text, we discern that Ross failed to make a favorable impression on Rachel, with whom he had a romantic interest. This understanding, combined with the extracted topic, indicates that his response carries an underlying sense of disappointment.

\section{Conclusion}\label{6}
We propose a Dynamic Causal Disentanglement model for emotion detection in dialogues. In our model, we introduce hidden variables to disentangle hidden variables and learn the causal relationships between utterances and emotions. We optimize our model by maximizing the ELBO. Experimental results and analysis on two popular datasets demonstrate the high accuracy and robustness of our model. In future research, we intend to incorporate more precise representations of dialogue as observational variables to further enhance the analysis of dialogues, with the expectation of achieving even greater performance.

\bibliography{cite}
\bibliographystyle{IEEEtran}

\vfill
\end{document}